\documentclass[conference]{IEEEtran}
\IEEEoverridecommandlockouts
\usepackage{cite}
\usepackage{amsmath,amssymb,amsfonts}
\usepackage{algorithmic}
\usepackage{graphicx}
\usepackage{textcomp}
\usepackage{xcolor}
\usepackage{diagbox}
\usepackage{subfigure}
\usepackage{mathtools}
\usepackage[section]{placeins}
\def\BibTeX{{\rm B\kern-.05em{\sc i\kern-.025em b}\kern-.08em
    T\kern-.1667em\lower.7ex\hbox{E}\kern-.125emX}}

\DeclareMathAlphabet{\pazocal}{OMS}{zplm}{m}{n}
\DeclareMathOperator*{\argmaxA}{arg\,max} 
\DeclareMathOperator*{\argmin}{arg\,min} 

\newcommand{\Lb}{\pazocal{L}}

\begin{document}

\title{Optimizing 3D Geometry Reconstruction from Implicit Neural Representations}

\author{\IEEEauthorblockN{1\textsuperscript{st} Shen Fan}
\IEEEauthorblockA{\textit{Dept. of Computer Science} \\
\textit{New Jersey Institute of Technology}\\
Newark NJ, USA \\
sf269@njit.edu}
\and
\IEEEauthorblockN{2\textsuperscript{nd} Przemyslaw Musialski}
\IEEEauthorblockA{\textit{Dept. of Computer Science} \\
\textit{New Jersey Institute of Technology}\\
Newark NJ, USA \\
przem@njit.edu}

}

\maketitle

\begin{abstract}
Implicit neural representations (INRs) have emerged as a powerful tool in learning 3D geometry, offering unparalleled advantages over conventional representations like mesh-based methods. A common type of INR implicitly encodes a shape’s boundary as the zero-level set of the learned continuous function and learns a mapping from a low-dimensional latent space to the space of all possible shapes represented by its signed distance function. However, most INRs struggle to retain high-frequency details, which are crucial for accurate geometric depiction, and they are computationally expensive. To address these limitations, we present a novel approach that both reduces computational expenses and enhances the capture of fine details. Our method integrates periodic activation functions, positional encodings, and normals into the neural network architecture. This integration significantly enhances the model's ability to learn the entire space of 3D shapes while preserving intricate details and sharp features, areas where conventional representations often fall short.
\end{abstract}


\section{Introduction}
Selecting the optimal method for representing 3D shapes and scenes has remained a crucial topic of effective 3D learning. Implicit neural representations (INRs) have emerged in the last few years and brought impressive advances to the state-of-the-art of learning-based 3D reconstruction and modeling, offering many possible benefits over conventional representations that use discrete structures such as voxels\cite{b1, b2}, meshes \cite{b3, b4}, and point clouds \cite{b5, b6}. INRs leverage on the possibility of deploying a multi-layer perceptron (MLP) to fit a continuous function that represents implicitly a signal of interest \cite{b7}. Characterized by continuous parameterization, INRs demonstrate a remarkable advantage in memory efficiency over discrete representations. This attribute grants them the ability to model fine detail which is free from the limitation of grid resolution and solely limited by the capacity of the underlying network architecture. They are able to describe the 3D geometry by fitting signed distance functions (SDF) \cite{b8, b10}, unsigned distance functions (UDF) \cite{b11, b12}, and occupancy fields (OCC) \cite{b13, b14}.

Neural SDF is a subset of this work and is one of the most effective approaches to represent 3D geometry. A signed distance function is a continuous function that, for a given spatial point, outputs the point’s distance to the closest surface, whose sign encodes whether the point is inside (negative) or outside (positive) of the watertight surface:
\begin{equation}
SDF(\boldsymbol{x}) = d, d\in \mathbb{R}, \boldsymbol{x}\in \mathbb{R}^3  \label{eq}
\end{equation}
The underlying surface is implicitly represented by the isosurface of SDF. A neural network with trainable parameters $\theta$ can be used to approximate the SDF to represent 3D shapes:
\begin{equation}
S = \{\boldsymbol{x} \in \mathbb{R}^3 | f(\boldsymbol{x};\theta)=0 \}\label{eq}
\end{equation}

Our work is inspired by DeepSDF \cite{b8}, which encodes all shapes in its training dataset into a latent space, allowing for smooth interpolation between different shapes. DeepSDF implicitly encodes a shape’s boundary as the zero-level set of the learned function while explicitly classifying space as either part of the shape’s interior or exterior. Despite its innovative approach, DeepSDF faces challenges in capturing the fine details of complex shapes. Like many recent INRs, DeepSDF relies on ReLU-based multi-layer perceptrons. However, due to the piecewise linear nature of ReLU networks, their second derivative is zero everywhere, making them incapable of capturing higher-order derivative information from input data.

To overcome this limitation, we propose incorporating periodic activation functions \cite{b15,b16} into our framework. Standard neural networks, especially simple architectures like MLPs, often struggle to learn high-frequency components due to spectral bias, which favors smoother, lower-frequency functions. In contrast, periodic activation functions are smooth and infinitely differentiable, making them ideal when the model output requires derivatives or when gradient calculations are involved. Their periodic nature enables implicit neural representations to capture high-frequency components more effectively.

However, we found that relying solely on periodic activation functions does not always result in high-quality reconstructions. To address this, we incorporate positional encoding methods \cite{b17,b18} to map coordinates into a higher-dimensional space, improving the model’s ability to capture sharp features. Additionally, we integrate normals into the learning process, providing essential geometric information that enhances surface detail and consistency, ultimately leading to more accurate and detailed shape reconstructions.

While most recent works \cite{b18,b19} successfully preserve sharp features, they are limited to representing a single shape. This paper aims to introduce a novel approach for learning the entire space of 3D shapes, effectively capturing sharp features. Additionally, it compares different periodic activation functions and positional encoding methods to determine their efficacy in this context. Our contribution includes that 
\begin{enumerate}
    \item A continuous implicit neural representation that learns the entire space of 3D geometric shapes while preserving sharp features.
    \item A simple yet effective paradigm for learning 3D shapes, enabling the reconstruction of smooth and detailed geometries by leveraging different periodic activation functions, positional encoding methods, and normal data. To the best of our knowledge, we are the first to evaluate the periodic activation function in \cite{b16}.
    \item A more robust model that achieves orders of magnitude improvements in both qualitative and quantitative comparisons with DeepSDF, while significantly reducing the training time required for scaling to larger models and datasets.
\end{enumerate}

\section{Related Works}
\subsection{Learning with Explicit Representations}

Explicit representations of 3D shapes can be broadly categorized into three main types: voxel-based, mesh-based, and point-based methods.

Voxel grids have traditionally been used to provide non-parametric representations of 3D shapes and voxel-based SDF representations have been extensively used for 3D shape learning \cite{b20, b21}.
However, these methods inherit the limitations of traditional voxel representations with respect to high memory requirements. 

To reduce the high memory cost, octree-based partitions are adopted \cite{b25, b2, b26}, relaxing the memory limitations of dense voxel methods and extending ability to learn at up to 512\textsuperscript{3} resolution. However, even this resolution is far from producing shapes that are visually compelling.
Mesh-based methods mostly deform a pre-defined mesh to approximate a given 3D shape. Various methods use existing \cite{b32, b33} or learned \cite{b34, b35} parameterization techniques to describe 3D
surfaces by morphing 2D planes. When using mesh representations, the quality depends on both the parameterization algorithms and the input mesh. There's a tradeoff between supporting arbitrary topology and achieving smooth, connected surfaces. To address this and generate closed mesh, works such as \cite{b32, b35} deform a sphere into more complex 3D shape, which produces smooth and connected shapes but limits the topology to shapes that are homeomorphic to the sphere.

Point clouds have achieved much attention recently due to its simplicity. Processing point clouds, however, is far from straightforward because of their unorganized nature. As a possible solution, some works projected the original point clouds to intermediate regular grid structures such as voxels \cite{b38} or images \cite{b39, b40}. Other methods like PointNet \cite{b5} proposed to operate directly on raw coordinates by means of shared multi-layer perceptrons followed by max pooling to aggregate point features. 

The limitation of learning with point clouds is that they do not describe topology and are not suitable for producing watertight surfaces.

\subsection{Implicit Neural Representations}
With the development of deep learning, INRs have achieved great progress in solving 3D geometry problems. In contrast to the explicit representations which require discretization, implicit models represent shapes continuously and naturally handle complicated shape topologies. INRs can broadly be categorized into three categories, namely global approaches, local approaches and meta learning.

Global implicit function approximation methods aim to represent all possible shapes by approximating a single continuous signed distance function. 
The surface of the object, defined as the zero-isosurface of the function, can then be efficiently extracted using algorithms like marching cubes \cite{b41}.
Some works, like \cite{b14}, propose assigning a value to each point in 3D space and using a binary classifier to extract an iso-surface. Occupancy networks \cite{b13} utilize a truncated SDF to determine the continuous boundary of 3D shapes. Unlike \cite{b14}, they predict the probability of occupancy within voxels, which can be used in a progressive multi-resolution process for refined outputs. SAL \cite{b42} parameterizes SDF with a neural network, learning the SDF by optimizing against the unsigned distance function, while SALD \cite{b43} extends this with supervision of derivatives.

Multiple latent codes are introduced in local approaches. The widely used strategy is to split the space occupied by the shape into a voxel grid. References such as \cite{b45, b48} decompose shapes into local patches, with each patch being represented by a corresponding latent code.
IF-Net \cite{b11} constructs hierarchical latent grids with different resolutions to capture local geometric information of different scales.

In contrast, DeepLS \cite{b49} similarly utilizes local representations from a 3D voxel grid, but it infers the latent codes within the voxel grid through auto-decoding. This approach significantly reduces training time compared to other methods, yet remains memory expensive. 

An alternative approach is gradient-based meta-learning. Recent algorithms aim to learn neural network initialization, which can be fine-tuned for a new task through a few steps of gradient descent \cite{b50,b51}. Implicit Model-Agnostic Meta-Learning (iMAML) \cite{b52} improves memory efficiency by obtaining gradients via an implicit method instead of backpropagating through unrolled inner loop iterations. Although meta-learning has not been widely explored for SDFs, MetaSDF \cite{b53} is the first to use gradient-based meta-learning for INRs but suffers from large time and memory complexity. Building on MetaSDF, GenSDF \cite{b54} introduces a two-stage semi-supervised meta-learning approach, generalizing well to unseen shapes without requiring prior information. However, it remains computationally expensive and does not achieve real-time inference due to the large number of model parameters.

\section{Methodology}

\subsection{Formulation}
We adopted the idea of auto-decoder from DeepSDF. Unlike the encoder-decoder networks, auto-decoder only has a decoder and directly accepts a latent vector as an input. We want to learn the latent space of shapes and for each category of N shapes we use auto-decoder $f_\theta$ to approximate the signed distance function ${SDF^i}_{i=1}^N$. For each shape, we prepare a set of M point samples and their signed distance values:
\begin{equation}
    X_i\coloneqq \{(\boldsymbol{x_j},d_j): SDF^i(\boldsymbol{x_j})=d_j\}_{j=1}^M \label{eq}
\end{equation}
Each latent code $\boldsymbol{z_i}$ is randomly initialized from a zero-mean Gaussian. The latent vectors are concatenated with sampled points as the input and are then jointly optimized during the training along with the parameters $\theta$.

Each shape in the given dataset is assumed following the joint distribution of shapes:
\begin{equation}
    p_\theta(X_i,\boldsymbol{z}_i) = p_\theta(X_i | \boldsymbol{z}_i)p(\boldsymbol{z}_i) \label{eq}
\end{equation}
where $\theta$ parameterizes the data likelihood. For a given $\theta$, a code $\boldsymbol{z}_i$ can be estimated via maximum-a-posterior (MAP) estimation:
\begin{equation}
    \boldsymbol{\hat{z}}_i = \argmaxA_{\boldsymbol{z}_i} p_\theta(\boldsymbol{z}_i | X_i) = \argmaxA_{\boldsymbol{z}_i} \log p_\theta(\boldsymbol{z}_i | X_i) \label{eq}
\end{equation}
$\theta$ can be estimated as the parameters that maximizes the posterior across all shapes:
\begin{equation}
    \begin{split}
        \boldsymbol{\hat{\theta}}_i &= \argmaxA_{\theta} \sum_{X_i \in \boldsymbol{X}} \max_{\boldsymbol{z}_i}  \log p_\theta(\boldsymbol{z}_i | X_i) \\
        & =\argmaxA_{\theta} \sum_{X_i \in \boldsymbol{X}} \max_{\boldsymbol{z}_i}  (\log p_\theta(X_i|\boldsymbol{z}_i)+\log p(\boldsymbol{z}_i)) \label{eq}
    \end{split}
\end{equation}

According to DeepSDF, $f_\theta$ is optimized by maximizing the joint log posterior over all training shapes with respect to the individual shape codes $\{z_i\}_{i=1}^N$ and the network parameters $\theta$. In practice, this is equivalent to minimize the deviation between the network prediction $\hat{d}_j$ from the actual SDF value $d_j$. The prior distribution over codes $p(\boldsymbol{z}_i)$ is assumed to be a zero-mean multivariate Gaussian with a spherical covariance $\sigma^2 I$. The final cost function becomes:
\begin{equation}
    \Lb_{loss} = \argmin_{\theta, \{z_i\}_{i=1}^N} \sum_{i=1}^N(\sum_{j=1}^M\Lb(f_\theta(\boldsymbol{x}_j,\boldsymbol{z}_i),d_j)+\lambda{\left \| \boldsymbol{z}_i \right \|}_2^2) \label{eq}
\end{equation}
where $\lambda = \frac{1}{\sigma^2} $ can be used to balance the reconstruction and regularization term and is set to $\lambda=10^{-4}$.  The loss function $\Lb$ is a $L_1$ loss function as follows:
\begin{equation}
    \Lb(f_\theta,d) = |clamp(f_\theta,\delta)-clamp(d,\delta)| \label{eq}
\end{equation}
where $clamp(x,\delta)$ constrains the value of $x$ within the range $[-\delta,\delta]$. If $x<-\delta$, it returns $x<-\delta$; if $x>\delta$, it returns $\delta$;otherwise, it returns $x$.

\subsection{Activation Functions}
The first periodic activation function we incorporate is derived from the sinusoidal representation networks or SIREN \cite{b15}, which are MLPs that utilize $sin(x)$ instead of ReLU as their activation function. Any derivative of a SIREN is a SIREN, as the derivative of the sine is a cosine, a phase-shifted sine. Therefore, the derivatives of a SIREN inherit the properties of SIRENs, enabling us to supervise any derivative of SIREN with complicated signals. SIREN demonstrates that training can be accelerated by introducing a frequency factor $\omega_0$ to $\sin(\omega_0 x)$ in all layers of the network and often set to a large value like 30. In our experiments, we also found that because we insert the frequency vector into all hidden layers, setting different values of $\omega_0$ for the first and subsequent layers can significantly affect the quality of the learned model.

The hyperbolic oscillation activation function (HOSC) \cite{b16}, which is another periodic parametric activation function used in our experiments, is defined as 
\begin{equation}
    HOSC(x;\beta)=tanh(\beta sin(x)) \label{eq}
\end{equation}
where $\beta$ is the sharpness parameter, enabling HOSC to seamlessly transition between a smooth sine like wave and a square signal. HOSC is differentiable not only with respect to the input x, but also with respect to the sharpness parameter $\beta$, which allows $\beta$ to be optimized during training. 

While the use of periodic activation functions enables our model to effectively capture and represent sharp features in 3D shapes, our experiments indicate that this new approach demonstrates increased sensitivity to hyperparameter settings and a greater tendency to overfit. To address this, we incorporated positional encoding methods to enhance the model's ability to capture sharp features more robustly.

\subsection{Positional Encoding}

To efficiently evaluate various positional encoding methods without increasing training time, we incorporated Tiny CUDA neural networks (tinycudann) \cite{b55}, a highly optimized C++/CUDA neural network framework with a PyTorch extension that supports a variety of positional encodings. After training on different shape categories, we found that although positional encodings enhance the model's ability to capture sharp features, not all methods are capable of fully learning the entire shape space. 
Among the encoding methods provided in tinycudann, we present reconstruction results using multiresolution hash encoding (HASH) \cite{b18} and fourier feature transform (FFT) \cite{b19}. While HASH performed well on individual objects, it struggled to learn the full space of shapes. We utilized the implementation of the FFT to present our best results.

\subsection{Normals}
Normals provide crucial geometric information that enhances surface detail and ensures greater consistency, leading to more accurate shape representations. Our research also demonstrates that incorporating normals during the shape learning process significantly improves the accuracy and quality of reconstructed models.

During 3D shape reconstruction training, incorporating normals in gradient calculations provides more informative gradients by capturing directional surface variations, allowing for more precise adjustments to model parameters.

We use normals as addtional regularization term to equation 3.7 and accurate normals can be analytically computed by calculating the spatial derivative $ \frac{\partial f_{\theta}(x)}{\partial x}$ via back-propagation through the network. We get the final loss function:
\begin{equation}
    \Lb_{final} = \Lb_{loss} + \tau{\left \| \nabla_xf_{\theta}(\boldsymbol{x},\boldsymbol{z}) - \boldsymbol{n} \right \|}_2^2) \label{eq}
\end{equation}
if normal data exists, $\tau = 1$, $\nabla_xf$ is close to the supplied normals.

\section{Evaluation Method}
Chamfer distance (CD) is a popular metric for evaluating shapes \cite{b56}. Given two point sets $S_1$ and $S_2$, the metric is simply the sum of the nearest neighbor distances for each point to the nearest point in the other point set.
\begin{equation}
    d_{CD}(S_1,S_2) = \sum_{x \in S_1} \min_{y \in S_2}{\left \| x - y \right \|}_2^2 + \sum_{y \in S_2} \min_{x \in S_1}{\left \| x - y \right \|}_2^2 \label{eq}
\end{equation}
To apply CD to meshes, we sample surface points and compute the CD on the resulting point clouds. In all of our experiments we report the CD for 30,000 points for both $|S_1|$ and $|S_2|$, which can be efficiently computed by use of a KD-tree.

\section{Results}

\subsection{Dataset}
We conducted our experiments on the ShapeNet dataset \cite{b57}, using the same data preparation method as DeepSDF. We trained our model across five categories: chairs, lamps, planes, sofas, and tables. Initially, we trained on a small subset of 100 objects to fine-tune the hyperparameters. However, these hyperparameters did not generalize well to larger datasets. To balance hyperparameter tuning time and reconstruction quality, we expanded our training to 1,000 objects for each of shape categories. Additionally, we found it necessary to make slight adjustments to the hyperparameters for different object categories to achieve optimal performance.

As shown in Table 4.1, when compared with DeepSDF, our proposed auto-decoder can reconstruct complex shapes with high frequency features, using less than $\frac{1}{3}$ of DeepSDF's parameters and training time.
\begin{table}[htbp]
    \caption{Comparison Between DeepSDF and proposed Auto-decoder}
    \centering
    \begin{tabular}{|c|c|c|c|c|} \hline
    \backslashbox{Methods}{Details} & Shapes & Parameters & Epochs & Training Time  \\ \hline
    DeepSDF & 1000 & 1.84M & 2000 & 30 hours\\ \hline
    Proposed & 1000 & 585K & 2000 & 12 hours\\ \hline
    \end{tabular}
    \label{tab:my_label}
\end{table}

\subsection{Qualitative Results}

\begin{figure*}[!htbp]
\centering
\includegraphics[width=.4\textwidth,height=.4\textheight,keepaspectratio]{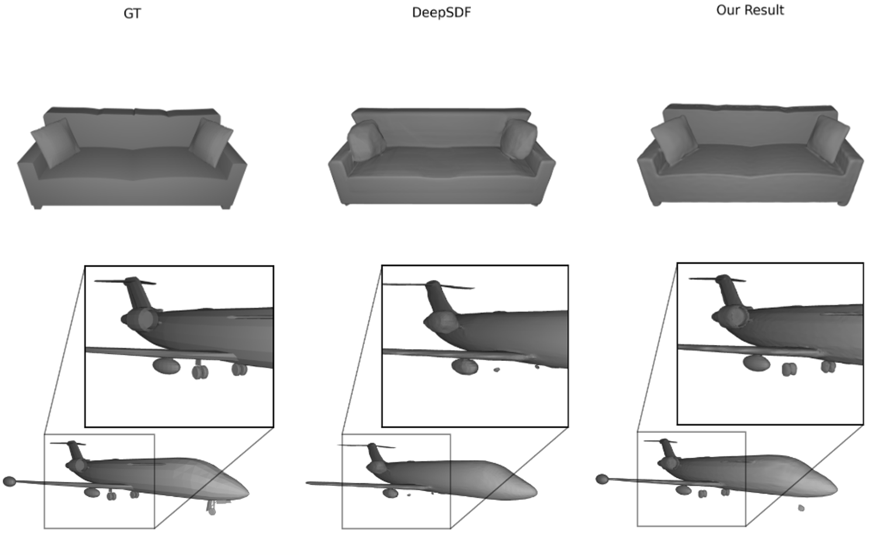}
\caption[Reconstruction Comparison Between DeepSDF and Our Model]{Reconstruction comparison between DeepSDF and our model.'GT' denotes the ground truth.}
\label{fig:2} 
\end{figure*}

Given that incorporating normals significantly enhanced the qualitative outcomes, all the qualitative and quantitative results presented in this paper, unless otherwise stated, are based on models that integrate normals. We begin by comparing our reconstruction results, using the SIREN activation function and FFT, with those from DeepSDF. As illustrated in Fig. 2, DeepSDF struggles to capture certain high-frequency details, such as back cushions, landing gears, and turbine inlets, whereas our model demonstrates superior performance in these areas. Nevertheless, our model still exhibits issues, such as unconnected landing gears in some plane reconstructions.

In Fig. 3, we compare the reconstruction results of models trained with HOSC and SIREN. While HOSC may introduce some noise, it captures sharp features more effectively than SIREN. However, tuning the hyperparameters of HOSC is more challenging, making it difficult to achieve the level of smoothness obtained with SIREN.
\begin{figure*}[!htbp]
\centering
\includegraphics[width=.4\textwidth,height=.4\textheight,keepaspectratio]{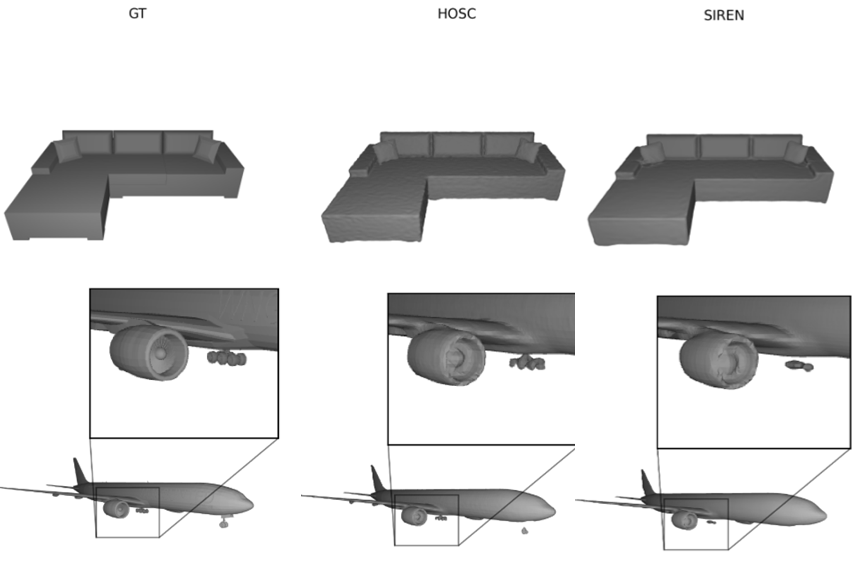}
\caption[Reconstruction Comparison Between DeepSDF and Our Model]{Reconstruction results using activation function from HOSC and SIREN, respectively.}
\label{fig:3} 
\end{figure*}

We evaluated various positional encodings and observed that, while all compatible positional encodings enhance our model's ability to capture finer details, not all are capable of learning the full shape space. As shown in Fig. 4, although the HASH encoding captures intricate details, it fails to accurately reconstruct one of the two sofas.

Fig. 5 shows that using normals with HOSC smooths defects and improves overall reconstruction quality. 
Fig. 6 highlights the robustness of our model in learning latent spaces across various shape categories; however, some shapes remain challenging to reconstruct accurately.

\begin{figure*}[!htbp]
\centering
\includegraphics[width=.4\textwidth,height=.4\textheight,keepaspectratio]{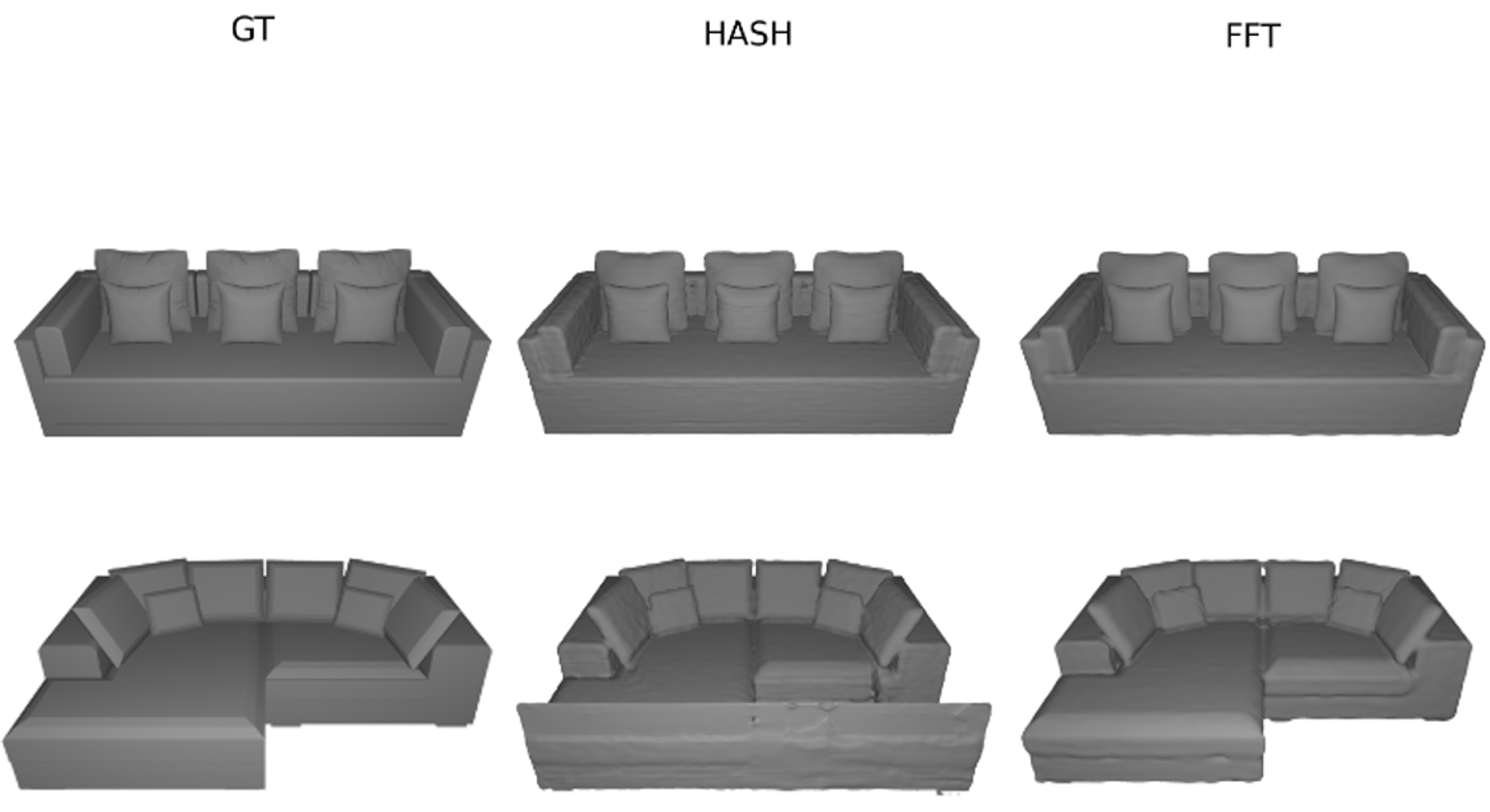}
\caption[Reconstruction Comparison Between DeepSDF and Our Model]{Reconstruction results using activation function from HOSC and SIREN, respectively..}
\label{fig:4} 
\end{figure*}

\begin{figure*}[!htbp]
\centering
\includegraphics[width=.4\textwidth,height=.4\textheight,keepaspectratio]{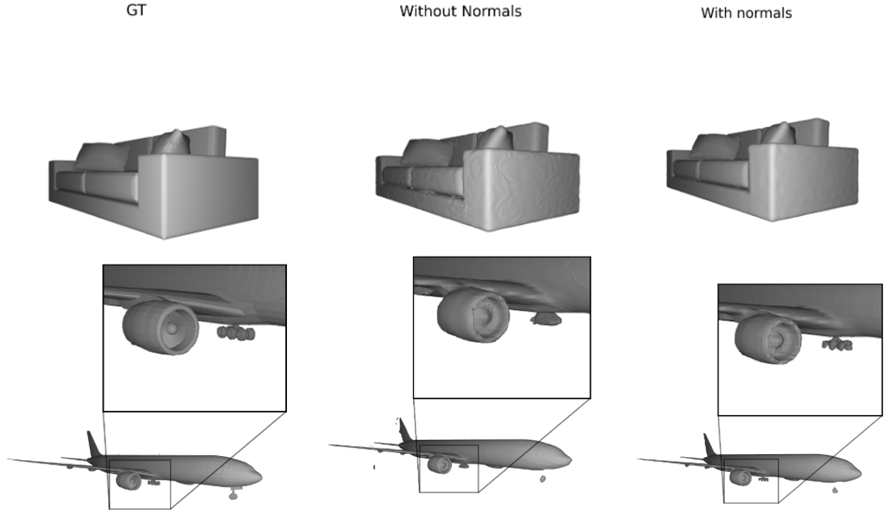}
\caption[Reconstruction Comparison Between DeepSDF and Our Model]{Reconstruction results with or without normals}
\label{fig:5} 
\end{figure*}

\begin{figure*}[!htbp]
\centering
\includegraphics[width=.5\textwidth,height=.5\textheight,keepaspectratio]{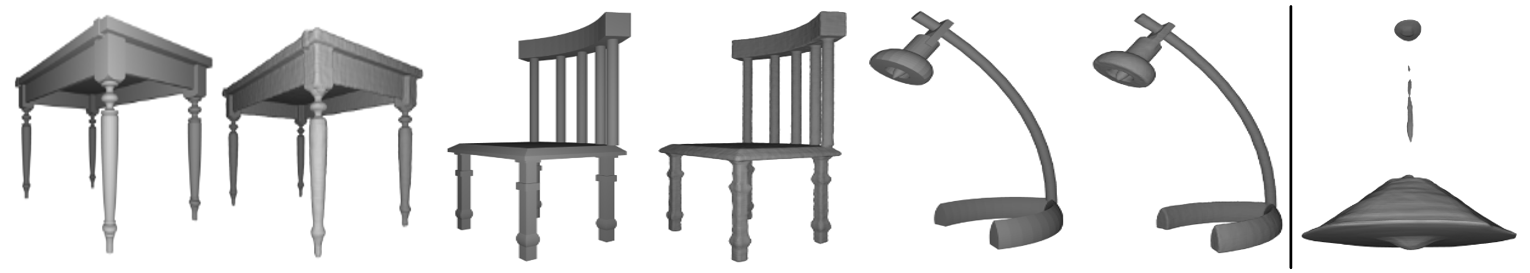}
\caption[Reconstruction Comparison Between DeepSDF and Our Model]{Reconstruction of test shapes. From left to right alternating: ground truth shape and our reconstruction. The right most column show failure case. This failure are likely due to lack of training data and failure of minimization convergence.}
\label{fig:6} 
\end{figure*}

\subsection{Quantitative Results}

We first show the quantitative result of model trained on small dataset (100 shapes). 
As shown in Table 4.2, using only a sinusoidal activation function, our method falls short of outperforming DeepSDF. However, with the addition of positional encoding, our proposed model, equipped with any periodic activation function, surpasses DeepSDF in terms of mean CD.

Table 4.3 presents the quantitative results for model trained on a large dataset. All models were trained on 1000 training data points for each category. Our approach, using only a periodic activation function, outperforms DeepSDF in median CD but shows a higher mean CD. This discrepancy is likely due to our network's tendency to overfit most shapes while neglecting outliers. Such behavior is further exacerbated by our use of an $L_1$-based loss function, which inherently encourages this pattern. To enhance our model's overall performance and address this issue, the development and integration of a new loss function are necessary.
With positional encoding, our proposed model can significantly reduce the CD, both in mean and median values. We only show the result with FFT because it is the best approach for our experiments.

\begin{table}[htbp]
    \caption{Quantitative Results of Model Trained on small dataset. CD = Chamfer Distance multiplied by $10^3$}
    \centering
    \begin{tabular}{|c|c|c|} \hline
    \backslashbox{Methods}{Metrics} & CD Mean & CD Median \\ \hline
    DeepSDF & 0.2076 & $\textbf{0.0432}$ \\ \hline
    SIREN & 0.2056 & 0.0875\\ \hline
    FFT/SIREN & 0.1284 & 0.0508\\ \hline
    FFT/HOSC & $\textbf{0.1142}$ & 0.0474\\ \hline
    \end{tabular}
    \label{tab:my_label}
\end{table}

\begin{table}[htbp]
    \caption{Quantitative Results of Model Trained on large dataset. CD = Chamfer Distance multiplied by $10^3$}
    \centering
    \begin{tabular}{|c|c|c|} \hline
    \backslashbox{Methods}{Metrics} & CD Mean & CD Median   \\ \hline
    DeepSDF & 0.1037 & 0.0422\\ \hline
    SIREN & 0.1541 & 0.0336\\ \hline
    FFT/SIREN & $\textbf{0.0493}$ & 0.0249 \\ \hline
    FFT/HOSC & 0.0523 & $\textbf{0.0197}$\\ \hline
    \end{tabular}
    \label{tab:my_label}
\end{table}

\section{Conclusion and Future Work}
In conclusion, periodic activation functions enhance the capacity for representing high-frequency features. However, we note that with only periodic activation functions, our model exhibits increased sensitivity to the settings of hyperparameters and tends to overfit. To address this, we incorporate normals, data augmentation, and positional encoding. Normals provide geometric stability and improve generalization, while positional encoding helps overcome the spectral bias of MLPs, making them better at learning high-frequency functions. Data augmentation diversifies the training set, mitigating overfitting.
While the incorporation of periodic activation functions, normals, and poisitional encodings into auto-decoder significantly enhances the preservation of fine details in 3D shape reconstruction, it also introduces challenges such as local INRs problems and unwatertight shapes for in the realm of implicit neural representations. Our future work will explore these challenges, aiming to balance the trade-offs between detail preservation, generalization to unseen shapes, and computational efficiency.

\end{document}